# Bilateral filters: what they can and cannot do

Oleg S. Pianykh, PhD

**Abstract**: Nonlinear bilateral filters (BF) deliver a fine blend of computational simplicity and blur-free denoising. However, little is known about their nature, noise-suppressing properties, and optimal choices of filter parameters. Our study is meant to fill this gap—explaining the underlying mechanism of bilateral filtering and providing the methodology for optimal filter selection. Practical application to CT image denoising is discussed to illustrate our results.

# Introduction

Efficient and accurate noise removal has always been in high demand with many quality-critical imaging applications, such as medical imaging. A cohort of image denoising techniques has been suggested in the literature, ranging from adaptive modifications of the linear filters to complex PDE-based iterative algorithms [Awate], [Tomasi], [Paris], [Jose], [Russo], [Phelippeau], [Tabuchi], [Gilboa], [Li], [Schilham]. *Bilateral Filters* (BF), originally developed in [Aurich], [Tomasi] and [Smith], seem to provide a nearly ideal combination of low computational complexity, flexibility, and original feature-preserving denoising.

Consider image pixel $p_0 = (x_0, y_0)$ and its neighbor $p = (x, y)$ with intensity values $f(p_0)$ and $f(p)$ respectively. Define two measures of pixel proximity:
1. $c(p, p_0)$ – spatial closeness, such as

$$c(p, p_0) = \exp\left(-\frac{1}{2}\left(\frac{\|p - p_0\|}{\sigma_d}\right)^2\right) = W(\|p - p_0\|) \qquad \text{(Eq.C)}$$

2. $s(f(p), f(p_0))$ – photometric similarity, such as

$$s(f(p), f(p_0)) = \exp\left(-\frac{1}{2}\left(\frac{\|f(p) - f(p_0)\|}{\sigma_r}\right)^2\right) = V(f(p) - f(p_0)) \qquad \text{(Eq.S)}$$

Then bilateral filtering function $H(f)$, transforming the original image $f$ into the filtered image $h = H(f)$, is defined at each pixel $p_0$ as

$$H(f)(p_0) = h(p_0) = \frac{\int_{p \in N(p_0)} f(p) c(p, p_0) s(f(p), f(p_0)) dp}{k(p_0)} =$$

$$= \frac{\int_{p \in N(p_0)} f(p) c(p, p_0) s(f(p), f(p_0)) dp}{\int_{p \in N(p_0)} c(p, p_0) s(f(p), f(p_0)) dp} \qquad \text{(Eq.Bil)}$$

where *N(p₀)* is the local *p₀* neighborhood (bilateral filter support), and the denominator *k(p₀)* normalizes the sum of weights to the unity. In other words, bilateral filtering averages neighboring *f(p)* with weights *c(p,p₀)s(f(p),f(p₀))* proportional to their Euclidean *c(p,p₀)* and photometric *s(f(p),f(p₀))* proximity to the central pixel *(p₀,f(p₀))*; pixels closest to *p₀* both in space and intensity contribute most to the filtered pixel value *h(p₀)*. This explains the principal advantage of BF over linear low-pass filtering: the use of nonlinear *c()* and *s()* functions preserves similar and close pixel values, avoiding blur in structural (non-random) image details. This attractive theory, however, leaves us with two principal questions:
1. The optimal choice of bilateral functions. Replacing *exp()* in (Eq.C) and (Eq.S) with any other positive decaying function will define another valid BF, but would it perform better?
2. The optimal choice of bilateral function parameters—such as $\sigma_d$ and $\sigma_r$ "spread" parameters in (Eq.C) and (Eq.S)

Simple image-filtering experiments [Tomasi] indicate that changes in function parameters alone can dramatically alter BF outcomes from excellent to completely unacceptable. Several approaches were used to deal with this problem: iterating the filter until it converges, estimating filter parameters with numerical models [Phelippeau], probabilistic analysis. However, all these techniques are not assumption-free, they take time to execute, and they cannot guarantee parameter optimality for the *given* image. Therefore, for real-life applications such as medical imaging, we need a straightforward, objective, and completely unsupervised BF optimization. Moreover—and most importantly—*we need to understand how bilateral filtering works*. Our study builds the necessary methodology to answer these questions.

## Defining parameters

### *Generic BF on a pixel grid*

Assuming one-dimensional *f(p)* (such as HU values for CT images), let us introduce positive decaying functions *V(x)* and *W(x)* such that

$$c(p, p_0) = W(\|p - p_0\|), \quad s(f(p), f(p_0)) = V(f(p) - f(p_0)),$$

$$W(0) = V(0) = 1 \quad \text{(normalization)},$$

$$W(\infty) = V(\infty) = 0 \quad \text{(vanishing)} \quad \text{(Eq.VW)}$$

$$V(x) = V(-x) \quad \text{(symmetry)}$$

$$W(x), V(x) \downarrow \quad \text{(decay)}$$

Functions *W* and *V* generalize exponents in (Eq.C) and (Eq.S). Then we rewrite (Eq.Bil) on discrete pixel grid as:

$$H(f)(p_0) = h(p_0) = \frac{1}{k(p_0)} \int_{p \in N(p_0)} f(p) c(p, p_0) s(f(p), f(p_0)) dp =$$

$$= \frac{\sum_r W(r) \left( \sum_{p \in R_r(p_0):\{\|p-p_0\|=r\}} f(p) V(f(p) - f(p_0)) \right)}{\sum_r W(r) \left( \sum_{p \in R_r(p_0):\{\|p-p_0\|=r\}} V(f(p) - f(p_0)) \right)} \quad \text{(Eq.BilR)}$$

Thus, the discrete bilateral filter summation in (Eq.BilR) separates into a *W(r)*-weighted average of range *V(f)* filters, taken on "rings" (fixed-radius domains) $R_r(p_0): \|p - p_0\| = r$ (Figure 1). The union of these rings defines the BF kernel support $N_n(p_0) = \bigcup_{i=1}^{n} R_{r_i}(p_0)$. This radial separation has one important advantage: even on relatively large supports, distance weights $w_i = W(r_i)$ comprise a relatively short sequence, which can be computed ahead of time for any known *W()*, greatly reducing the computational cost of bilateral filtering. Therefore, instead of *W()* one can study a small set of "radial weighting" parameters $w_i = W(r_i)$, defined for pixel grid radii $r_i = 1$, $\sqrt{2} = \sqrt{1^2 + 1^2}$, $2 = \sqrt{2^2 + 0^2}$, $\sqrt{5} = \sqrt{2^2 + 1^2}$, and so on. Moreover, as $r_i$ increases:
1. The difference $|r_i - r_{i+1}|$ becomes smaller, suggesting representing close $r_j$ with the same weight $w_i$,
2. The gain in image denoising from large $r_i$ becomes negligible.

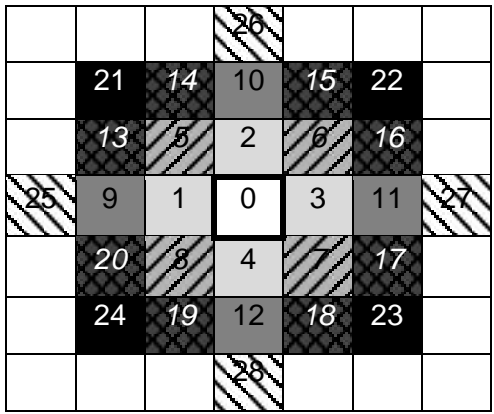

Figure 1: BF pixel neighborhood; shading corresponds to different rings $R_i$, and pixel numbers show "radial" pixel indexing we used.

This suggests (and will be demonstrated later), that BF can be implemented with a very few radial weights $w_i$. In our numerical experiments, we used up to *n* = 5 domain rings $R_i$ (shown with different shading in Figure 1).

Conversely, pixel differences *f(p)-f($p_0$)* correspond to much more varying and subtle choices than BF kernel radii, and *V(f(p)-f($p_0$))* cannot be reduced to a limited set of parameters[1]. But in addition to exponential *V()* in (Eq.S) we need to consider other *V()* satisfying (Eq.VW), as candidates for better filter properties.

## *BF as deviation-reducing transform*

Bilateral filters *H(f)* possess one important quality that easily follows from (Eq.Bil): they are shift-invariant in the intensity domain. This means that for any constant $f_0$, and on any filter support:

*H(f(p)+$f_0$) = H(f(p))+$f_0$ = h(p)+$f_0$*   (Eq.Shift)

Let's define $f_0$ as the average $f_a$ of *f(p)* over the local filter support *N($p_0$)*:

---

[1] This uncovers another BF property: bilateral filters are more sensitive to the changes in pixel values, than to the changes in pixel distances. One way to overcome it would be extending BF to subpixel level, but we leave this outside the scope of our current analysis.

$$f_a = \frac{1}{|N(p_0)|} \sum_{p \in N(p_0)} f(p)$$

then we can limit our study to zero-averaged $f_a(p) = f(p)-f_a$ after applying the $f_a$ shift[2]. This makes the deviation of $f(p)$ over $N(p_0)$

$$t = std(f_a(p)) = std(f(p)) = \sqrt{\frac{1}{|N(p_0)|} \sum_{p \in N(p_0)} (f(p) - f_a)^2}$$

the next and the most important factor, affecting BF outcomes.

What factors contribute to this deviation? Due to the limited size of $N(p_0)$, the variance in $f(p)$ over $N(p_0)$ cannot be attributed to the low-frequency components of the image. Instead, it originates from two primary high-frequency sources:
1. Unstructured $f(p)$ changes: random noise
2. Structured $f(p)$ changes: sharp local image details (edges)

It is expected that random changes have smaller amplitude compared to the structured image details—otherwise they would have been undistinguishable. Our goal is to reduce the local random noise deviation, while preserving larger intensity changes associated with structured image details. In other words, given local zero-average intensity $f_a(p) = f(p)-f_a$, we want to construct a filter that works selectively on different deviation "bands" in $f_a(p)$.

Linear filters do not possess this quality: they uniformly reduce $f(p)$ deviation $std(f)$ by a constant factor (corresponding to the linear filter coefficients). This results in blurred image details, and distorted local patterns. We expect that nonlinear BF can produce non-constant deviation changes. Therefore, we want BF to reduce small-scale deviations (noise) to preserve large-scale (structural) edges and to provide a smooth transition in between (to avoid sharp deviation thresholding with visible artifacts). The optimal $W()$ and $V()$ in (Eq.BilR) should be chosen accordingly.

This reasoning can be expressed in the frequency domain, but we prefer to study BF as the local deviation transforms, which makes it easier to interpret and apply. We can view $f_a(p) = f(p)-f_a$ as a random process $t \times Q$ where random $Q$ has zero mean and unit deviation, and scaling factor $t$ represents the local deviation in $f_a(p)$.

As a result, the main direction of our analysis will be evaluating BF for different $t$ and different types of random distribution $Q$.

## Parameter optimality

We evaluated BF behavior on $n = 5$ expanding ring-like neighborhoods (Figure 1): pixel ring $R_1$ formed by pixels 1–4, $R_2$ formed by 5–8, $R_3$ by 9–12, $R_4$ by 13–20, and $R_5$ by 21–28. Consequently, the five BF support areas were $N_1 = R_1$, $N_2 = R_1+R_2$, ..., $N_5 = R_1+R_2+R_3+R_4+R_5$. Note that in the case of $R_5$ we combined two "subrings" formed by pixels 21–24 and 25–28, with very close Euclidean radii of $2\sqrt{2}$ and 3. Larger supports ($N_k$, $k > 5$) were also tested, but the improvements in the filter quality were found negligible and not worth the additional computational expenses.

---

[2] Shift-invariance is another BF limitation: they cannot be effectively applied to the images where the local noise depends on the bias $f_0$.

## Role of filter functions V() and W()

We experimented with power $w_i = 2^{-i}$ ($W(x) = 2^{-x}$) and exponential $w_i = e^{-(i/r_i)^2}$ ($W(x) = e^{-(x/r_i)^2}$) functions as rather standard choices for the radial weighting $W(x)$; and we used the following $V(x)$ with variance-scaling parameter $t$:

Abs: $$V_{abs}(x) = \frac{1}{1+|tx|}$$ (Eq.VA)

Frac: $$V_{frac}(x) = \frac{1}{1+(tx)^2}$$ (Eq.VF)

Quad: $$V_{quad}(x) = \frac{1}{1+(tx)^4}$$ (Eq.VQ)

Exp: $$V_{exp}(x) = e^{-(tx)^2}$$ (Eq.VE)

For each choice of scale $t$ and support size $n$, the local changes in $f(p)$ were modeled with two random distributions $Q$: normal and uniform. Normal distribution seems to provide the best practical match to the local CT noise (as we will see below); and uniform distribution was used in contrast with the normal to better reveal the effects of the distribution type over the BF performance.

To measure deviation reduction as a function of BF parameter selection, we defined and studied the following *deviation reduction function*:

$$K(t,n) = \frac{std_{p_0}(H(f))}{std_{p_0}(f)}$$ (Eq.K)

In this equation, $std_{p_0}$ defines the standard deviation $std(f_a(p)) = std(f(p)-f_a) = std(f(p))$ in the neighborhood of the pixel $p_0$. In other words, for any choice of bilateral filter parameters $(t,n)$ and functions $V()$ and $W()$, BF deviation reduction function $K(t,n)$ represents the ratio of the local deviation in the filtered image $h = H(f)$ to the local deviation of the original image $f$. The typical $K(t,n)$ curves as functions of $t = std(f_a(p))$ are shown in Figure 2—corresponding to Abs, Frac, Quad, and Exp choices of $V()$, $W(x) = 2^{-x}$, $n = 3$, (12-pixel support $N_3$, pixels from 1–12 in Figure 1), and normally-distributed[3] $Q$:

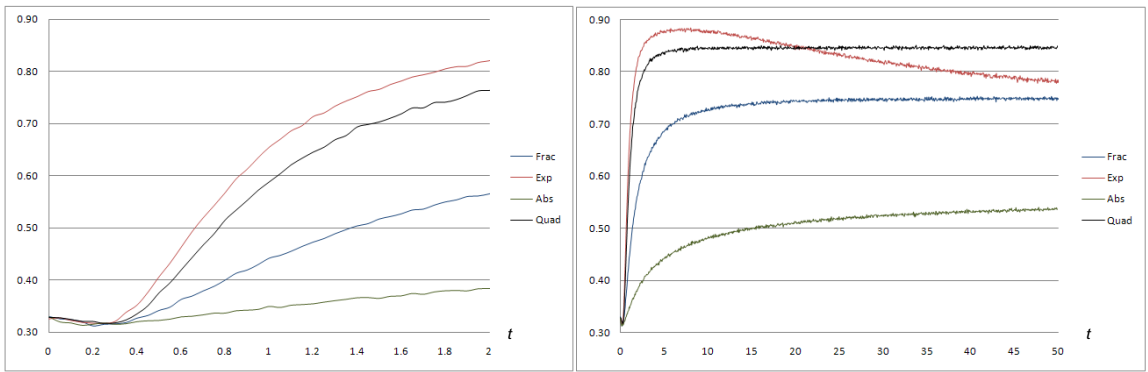

Figure 2: $K(t)$ for four different choices of $V()$ in (Eq.VA)-(Eq.VE), normally-distributed $Q$, 12-pixel support $N_3$. The left graph details the segment $0 \leq t \leq 2$, where $K(t)$ has local minimum. The right graph clearly shows the decay in the exponential filter (Exp) after $t = 5$.

The graphs in Figure 2 already reveal many interesting properties of bilateral filters:

---

[3] All $K(t,n)$ curves were computed numerically, hence small irregularities in their shapes.

1. Bilateral filters do work as local deviation thresholders: they substantially smooth low-deviation signals ($t < 1$), and have a lesser effect on high-deviation ones. The thresholding is "fuzzy," and cannot be associated with any sharp cutoff value. Therefore, we suggest defining BF threshold $t_0$ as $K(t_0) = 0.5$ (local deviation is reduced by half), and for our choices of $V()$ this amounts to the values shown in Table 1.

   Table 1: 50% deviation reduction thresholds for different choices of $V()$

   | Choice of $V()$ | Abs | Frac | Quad | Exp |
   | --- | --- | --- | --- | --- |
   | $t_0$ | 15.05 | 1.40 | 0.77 | 0.67 |

   As you can see, the value of $t_0$ is greatly affected by the choice of $V()$, and can be very far from 1.0 for slow-decaying $V()$.

2. $K(t) < 1$ means that BF will always reduce the local deviation of the original signal—as expected, bilateral filters are intrinsically smoothing. It is the *degree of smoothing* that is governed by the choice of BF parameters.

3. $K(0)$ corresponds to BF acting as the linear low-pass filter $W()$ with $w_i = W(r_i)$ coefficients. Therefore, it does not depend on the BF function $V()$ or data distribution $Q$, and for $t$ close to 0 we can apply well-known linear filter theory—in particular, for selecting the optimal function $W()$.

4. For $t > 0$, $K(t)$ has local minimum $K_{min} = K(t_{min}) < K(0)$ (see $t \approx 0.2$ in Figure 2, left). This means that for small $t$ bilateral filters are not guaranteed to perform better than linear low-pass filters $W()$ they correspond to.

5. $K(t)$ may have local maxima, and this clearly depends on the filtering function $V()$. As one can see from Figure 2, $K(t)$ for $V_{exp}()$ in (Eq.VE) decays after $t = 10$, thus worsening the performance of the corresponding bilateral filter (blurring high-magnitude image details). In comparison, $K(t)$ for the other functions remains nearly constant. The advantage of the $V_{exp}()$ is in the higher $K(t)$ maximum ($K_{max} = 0.8838$), which means better preservation of the high-deviation details. The disadvantage is that this maximum is narrow and can be easily missed in practical applications. Therefore, the exponent $V_{exp}()$ in (Eq.VE) should be viewed as the least robust choice.

6. Out of Frac, Abs, and Quad $V()$ choices the Quad provides the best high-deviation preservation for large $t$ (higher $K(t)$). However, this also corresponds to a sharper $K(t)$ increase after $t_0$, which can result in visible artifacts when preserved high-deviation details become surrounded by artificially-looking blur from the removed low-deviation noise. Therefore we chose Frac $V()$ as a reasonable compromise between slower increase around $t = t_0$ and higher $K(t)$ for $t > t_0$.

## *Role of filter support size n*

The role of the filter support size *n* is illustrated in Figure 3, showing $K(t,n)$ for $n = 1$ ($N_1$, 4 pixels), $n = 2$ ($N_2$, 8 pixels), $n = 3$ ($N_3$, 12 pixels), $n = 4$ ($N_4$, 20 pixels), and $n = 5$ ($N_5$, 28 pixels):

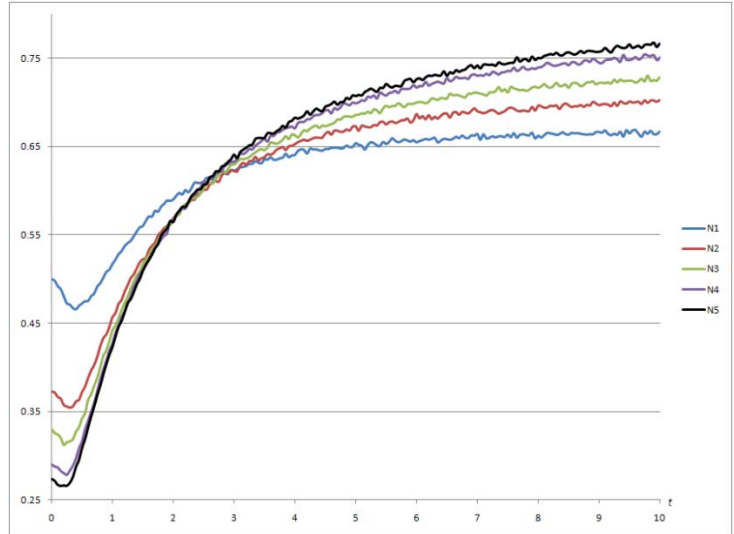

**Figure 3: Role of BF support size n. Normal Q, V() as in (Eq.VF), $W(x) = 2^{-x}$**

As one would expect, increasing BF support size *n* generally serves to the filter advantage, providing more noise suppression in the low-deviation range $t < t_0$, and more detail preservation for high-deviation areas (large *t*). However, for large *n,* this gain becomes considerably smaller (compare $N_4$ and $N_5$ curves in Figure 3), yet comes at a high computational cost (proportional to $n^2$). For that reason, we limited our optimal filters to $n = 3$ ($N_3$, 12 pixels) and $n = 4$ ($N_4$, 20 pixels).

For some V(), BF support size *n* can play yet another role. Figure 4 shows the changes in $K(t,n)$ as *n* varies from 1 to 5, and V() is chosen as in (Eq.VE). Similarly to Figure 2, $K(t,n)$ in Figure 4 becomes more "thresholding" for higher *n*, but it also exhibits a larger decay after its local maximum. This can be viewed as another argument against exponential V() in bilateral filters.

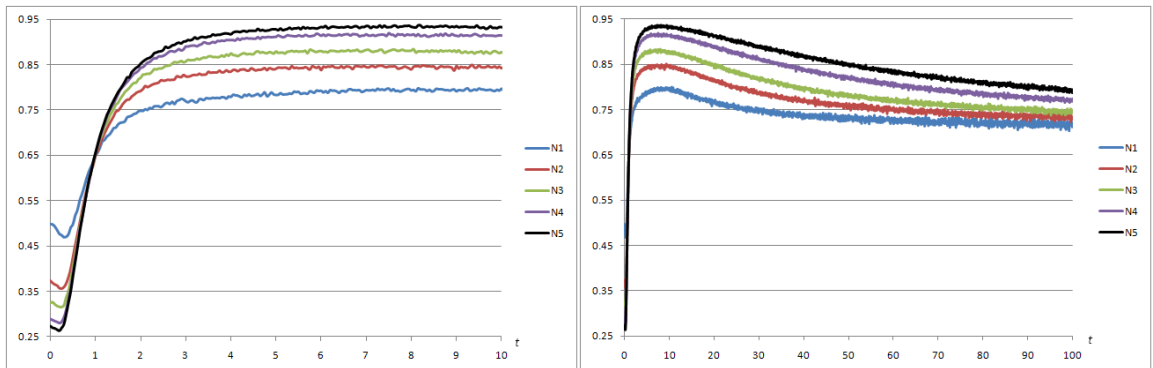

**Figure 4: Role of BF support size n for Exp choice of V(), (Eq.VE).**

## The role of distribution Q

We used random distribution *Q* to describe the local changes in the pixel intensities. Figure 5 compares Frac (Eq.VF) and Exp (Eq.VE) choices of *V*() applied to the normal and uniform *Q* (*n* = 3):

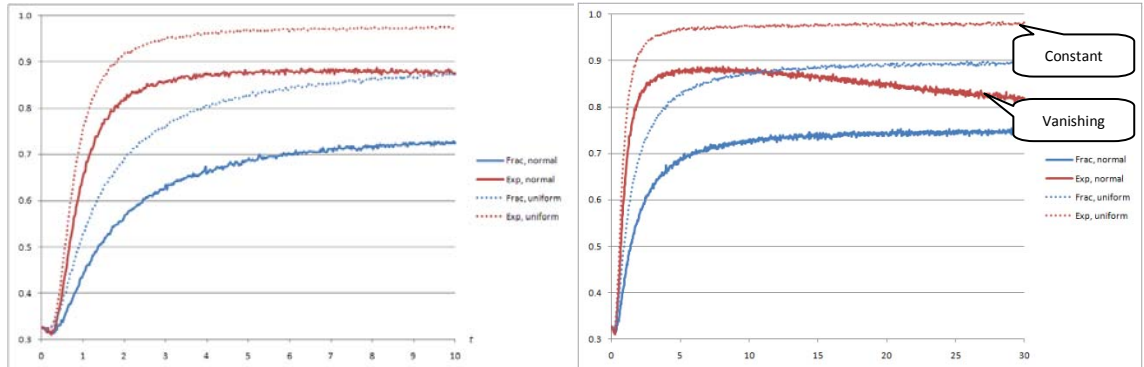

Figure 5: K(t) curves for Exp and Frac choices of *V*() on $N_3$ support. Note that Exp *V*() for uniform *Q* (dotted red line) does not have the local maximum as for normal *Q* (solid red line)

As one would anticipate, the choice of *Q* matters and can greatly affect the bilateral filter performance, especially for large $t > t_0$. Bilateral filters will be more smoothing for the normally-distributed *Q* (lower values of *K(t)*). Moreover (Figure 5, right), for large *t*, *K(t)* for $V_{exp}$() remains constant on uniform *Q*—contrary to the vanishing pattern for normal *Q*. Therefore, correct estimates of the local noise distribution type are essential to BF applications.

# Illustration: Practical application to CT radiation reduction

We applied our analysis to build an optimal bilateral filter to enhance low-radiation CT images. In CT imaging, increasing image quality leads to a thorny dilemma: visualization of the most subtle image details exposes patients to higher X-ray doses. Conversely, low radiation imaging is associated with a number of quality-degrading artifacts, the most prominent of them being noise. Noise obscures diagnostically-valuable details; and if it can be removed with some robust image post-processing technique, lower radiation scans become possible.

This reasoning has become the *raison d'être* for various CT noise-reducing filters, usually applied in two domains:
1. Filtering noise in the raw CT data—before it is rasterized to DICOM images.
2. Filtering noise in DICOM images.

The first approach works with the originally-acquired data, and provides more control over quality image denoising. However, it entirely depends on the CT equipment and therefore can be explored by CT vendors only. For instance, General Electric's Adaptive Statistical Iterative Reconstruction (ASIR) technique reduces noise and improves contrast in raw CT data before it is converted to DICOM images [ASIR].

The second approach discussed in this manuscript is vendor independent and therefore applicable to any digital images. Unfortunately, it deals with data that has already been subsampled and rasterized, leaving us with fewer noise-reduction options. Nonetheless, it is more straightforward and universal, which makes it particularly attractive in multi-vendor clinical environments. Can this be done with bilateral filters?

## *Low-radiation CT noise*

First of all, one would need to demonstrate that low-radiation noise in CT images possesses properties that can be targeted with bilateral filters.

To study noise distribution at different radiation levels, we used a plastic phantom shown in Figure 6 (left). The phantom was scanned at 100, 75, 50, and 25 percent of the regular liver protocol CT current (120kV, 200mA, exposure 500msec), and noise distribution was analyzed at each scan for each phantom density. Figure 6 (right) shows an example of noise distribution histograms, computed at different phantom densities, and shifted to zero mean for comparison. As one can conclude, all noise distributions turned out to be normal ($p < 0.001$) and nearly identical to each other (same standard deviation within 5% error margin). This proved that:
1. Shift-invariant bilateral filters can be indeed applied to CT noise data, as CT noise is shift-invariant as well.
2. Normal distribution $Q$ is an appropriate model for CT radiation noise.

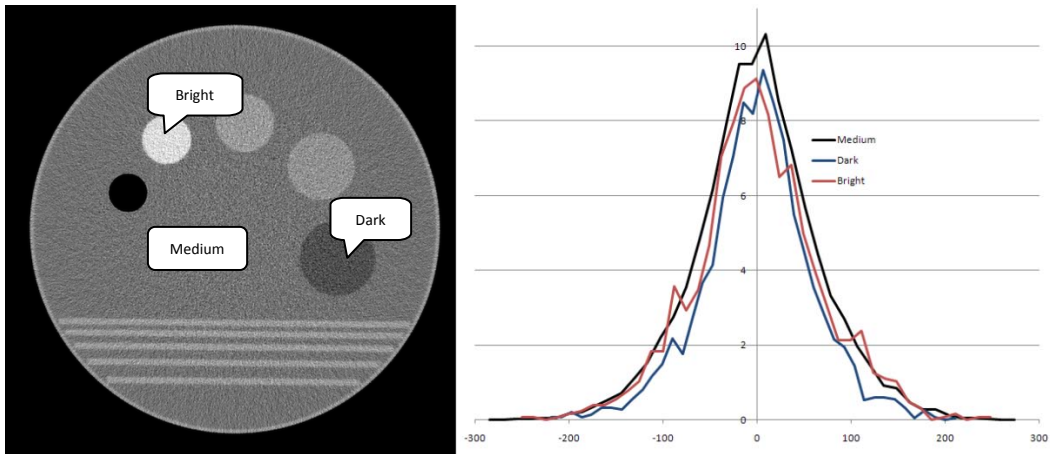

Figure 6: CT phantom (left) and noise histograms for the three phantom areas, shifted to the same zero mean for comparison. The original mean densities were 2.24HU (Medium), -108HU (Dark) and 334HU (Bright).

This conclusion was confirmed with linear regression analysis, where local deviation $s_H = std_{p_0}$, measured at each $p_0$ in a high-radiation image, was regressed on corresponding local deviation $s_L = std_{p_0}$ and local intensity averages $f_a$ in low-radiation images. While the p-values for $s_L$ were significant ($p < 1.0e-10$), those for $f_a$ did not show any statistical dependency ($p = 0.9$), suggesting that low-radiation CT noise can indeed be viewed as shift-invariant (independent of the average intensity). Finally, we computed the empirical $K(t) = s_H/s_L$, as shown on Figure 7, fitting it with a smooth trend polynomial. The polynomial provided a very significant fit ($R^2 = 0.6798$), with the polynomial shape being very close to what we have seen on our BF analysis graphs (Figure 2 - Figure 5).

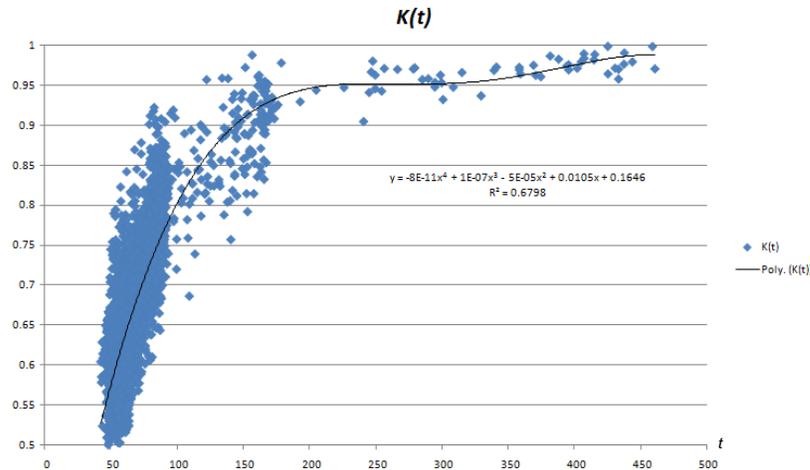

Figure 7: Empirical *K*(*t*), derived from CT phantom data, corresponding to 50% radiation reduction.

These practical observations support the applicability of bilateral filters to low-radiation CT denoising[4].

## *Bilateral filters for CT denoising*

In a typical clinical environment, CT images are acquired with preset CT scanning protocols—scanner parameter sets, fine-tuned for each organ. The main advantage of these protocols is consistency—essential for diagnostic image comparison and interpretation. Therefore, we applied our method on the CT protocol level using the liver scanning protocol mentioned above, and selected protocol-optimal BF settings to support lower CT currents (radiation).

Based on the phantom analysis, we discovered how local noise deviation $s = std_{p_0}$ depends on the CT dose reduction fraction *x* (Figure 8):

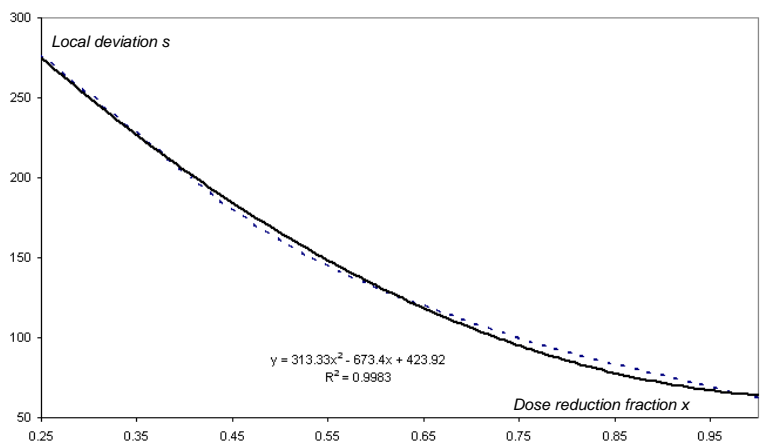

Figure 8: Local deviation as a function of CT dose reduction (polynomial trend match is shown with solid line).

---

[4] We are well aware of more complex aspects of CT noise, such as dependence on the direction and location (mostly due to photon starvation effects). However, these effects will be present in any CT image, noisy or noise-free; they are intrinsic to CT imaging. Therefore, we leave these aspects outside the scope of our study.

The trend of this change was closely approximated as $s = 313.33 x^2 - 673.4x + 423.9$ ($R^2 = 0.9983$), where $x$ represents the CT dose reduction fraction (varying from 0.25 to 1.0 of the full CT current), and $s$—local pixel deviation, attributed to noise. This resulted in the following optimal filter parameter strategy:

1. Find image dose reduction $x$ from the image scan parameters, as a fraction of the full-dose CT current[5].
2. Approximate low-radiation noise deviation as $s = 313.33 x^2 - 673.4x + 423.9$
3. Set optimal BF scaling parameter $t$ to achieve the desired percentage of noise reduction. We used 50% reduction, therefore we set $t = t_0/s$ (where 50% noise reduction threshold $t_0$ is given in Table 1).

As described earlier, we preferred $V_{frac}()$ and $n = 3$ as the filter/function selection, thus $t_0 = 1.40$ was used as the 50% deviation-reducing threshold (Table 1, Figure 2). We also limited filtering to the [-100;300] HU range, to exclude background and non-diagnostic structures[6], and to achieve faster performance. The result of our BF-optimal filtering was compared to the scanner-side GE ASIR filter: 4 radiologists visually graded 18 CT images, filtered with our BF-optimal and GE ASIR filters, on the scale from 1 to 5 (5 being the diagnostically-best image appearance). CT images with lung and liver lesions were used. Table 2 lists the averaged visual quality scores found.

Table 2: Visual comparison of our BF-optimal filtering with raw-data ASIR filters

| Filter type | ASIR 20% | ASIR 50% | Our BF-optimal 50% |
|---|---|---|---|
| Visual quality score | 2.3 | 2.7 | 3.5 |

This was considered to be a practical validation of our optimal filter approach.

As mentioned earlier, one of the major advantages of bilateral filtering is its computational simplicity, allowing for real-time applications. We implemented our filtering in C++ on a rather moderate 2.5GHz PC, and achieved a processing speed of 4–5 CT images/second (including loading and parsing DICOM image files). This speed is well-compatible with our PACS image retrieval speed; and because these two tasks can be done in parallel (filtering retrieved images while retrieving the next), the burden of filtering on the clinical workflow becomes unperceivable. There is no doubt that implementing BF with optimal filtering directly on the PACS workstations with more powerful hardware will reduce computational overhead to practically nothing.

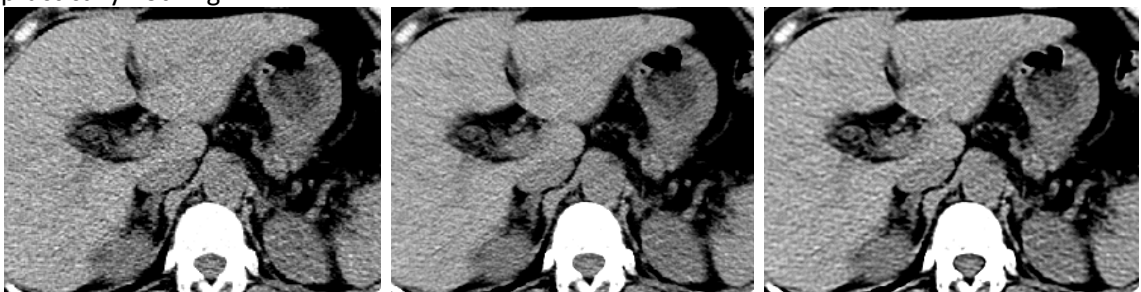

Figure 9: Original image (left), and the same image after applying GE ASIR-50 (middle), and our optimal (right) filters.

Will repetitive (iterative) BF application work even better? Not really. Bilateral filters, as follows from our analysis, are intrinsically smoothing, and will flatten even the most prominent image details when applied repetitively: $K(K(K(…K(t))))\to 0$. This is illustrated in Figure 9: note the considerable increase in flat-intensity areas on the rightmost image.

---

[5] Can be easily extracted from DICOM image tags.
[6] We worked with non-contrast CT of the liver.

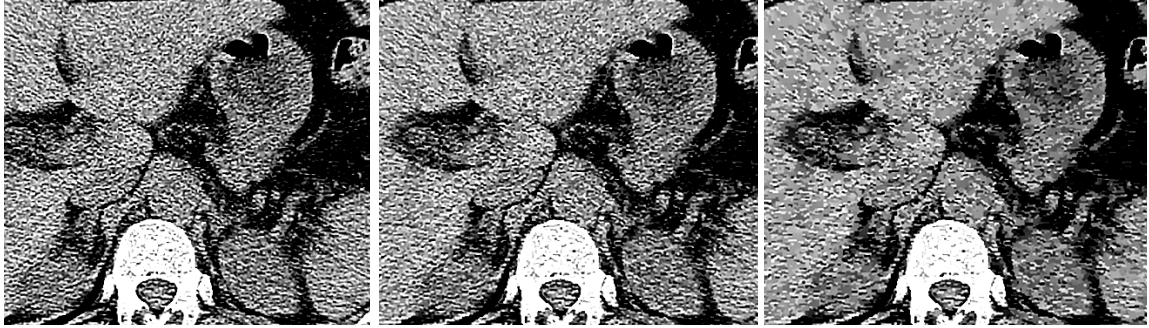

Figure 10: Iterative BF artifacts: original image (left), 5 BF iterations (middle), 10 BF iterations (right).

## Conclusions

We presented a new methodology for optimal BF parameter selection, based on $K(t)$ noise reduction function analysis, and illustrated it with CT image denoising. Our approach is completely unsupervised and simple enough to be implemented efficiently in a clinical environment. As future research, we see large potential in further $K(t)$ curve analysis to define nonlinear $V()$ and $W()$ with application-optimal properties.